\documentclass[10pt,twocolumn,letterpaper]{article}

\usepackage{cvpr}
\usepackage{times}
\usepackage{epsfig}
\usepackage{graphicx}
\usepackage{amsmath}
\usepackage{amssymb}

\usepackage{multirow}
\usepackage[breaklinks=true,bookmarks=false]{hyperref}

\cvprfinalcopy 

\def\cvprPaperID{****} 

\begin{document}

\title{Stacked Conditional Generative Adversarial Networks for Jointly Learning Shadow Detection and Shadow Removal}
\author{Jifeng Wang \thanks{{co-first author}}
	, Xiang Li$^*$, Le Hui, Jian Yang\\
	Nanjing University of Science and Technology\\
	{\tt\small jfwang.cs@gmail.com, \{xiang.li.implus, le.hui, csjyang\}@njust.edu.cn}
}

\maketitle

\begin{abstract}
   Understanding shadows from a single image spontaneously derives into two types of task in previous studies, containing shadow detection and shadow removal. In this paper, we present a multi-task perspective, which is not embraced by any existing work, to jointly learn both detection and removal in an end-to-end fashion that aims at enjoying the mutually improved benefits from each other. Our framework is based on a novel STacked Conditional Generative Adversarial Network (ST-CGAN), which is composed of two stacked CGANs, each with a generator and a discriminator. Specifically, a shadow image is fed into the first generator which produces a shadow detection mask. That shadow image, concatenated with its predicted mask, goes through the second generator in order to recover its shadow-free image consequently. In addition, the two corresponding discriminators are very likely to model higher level relationships and global scene characteristics for the detected shadow region and reconstruction via removing shadows, respectively. More importantly, for multi-task learning, our design of stacked paradigm provides a novel view which is notably different from the commonly used one as the multi-branch version. To fully evaluate the performance of our proposed framework, we construct the first large-scale benchmark with 1870 image triplets (shadow image, shadow mask image, and shadow-free image) under 135 scenes. Extensive experimental results consistently show the advantages of ST-CGAN over several representative state-of-the-art methods on two large-scale publicly available datasets and our newly released one.
\end{abstract}

\section{Introduction}

Both shadow detection and shadow removal reveal their respective advantages for scene understanding. The accurate recognition of shadow area (i.e., shadow detection) provides adequate clues about the light sources \cite{lalonde2009estimating}, illumination conditions \cite{panagopoulos2009robust,panagopoulos2011illumination,panagopoulos2013simultaneous}, object shapes \cite{okabe2009attached} and geometry information \cite{junejo2008estimating,karsch2011rendering}. Meanwhile, removing the presence of shadows (i.e., shadow removal) in images is of great interest for the downstream computer vision tasks, such as efficient object detection and tracking \cite{cucchiara2001improving,mikic2000moving}. Till this end, existing researches basically obey one of the following pipelines for understanding shadows:

\begin{figure}[t]
	\begin{center}
		\setlength{\fboxrule}{0pt}
		\fbox{\includegraphics[width=0.48\textwidth]{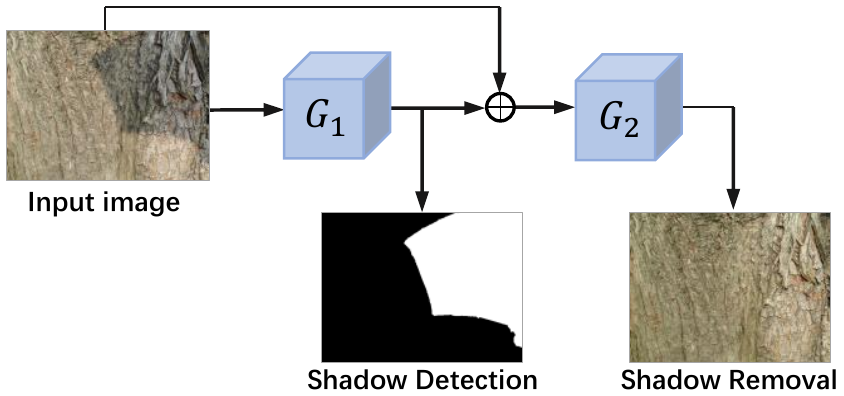}}
	\end{center}
	\vspace{-10pt}
	\caption{We propose an end-to-end stacked joint learning architecture for two tasks: shadow detection and shadow removal.}
	\label{fig_tasks_cropped}
	\vspace{-12pt}
\end{figure}

\begin{figure*}[t]
	\begin{center}
		\setlength{\fboxrule}{0pt}
		\fbox{\includegraphics[width=\textwidth]{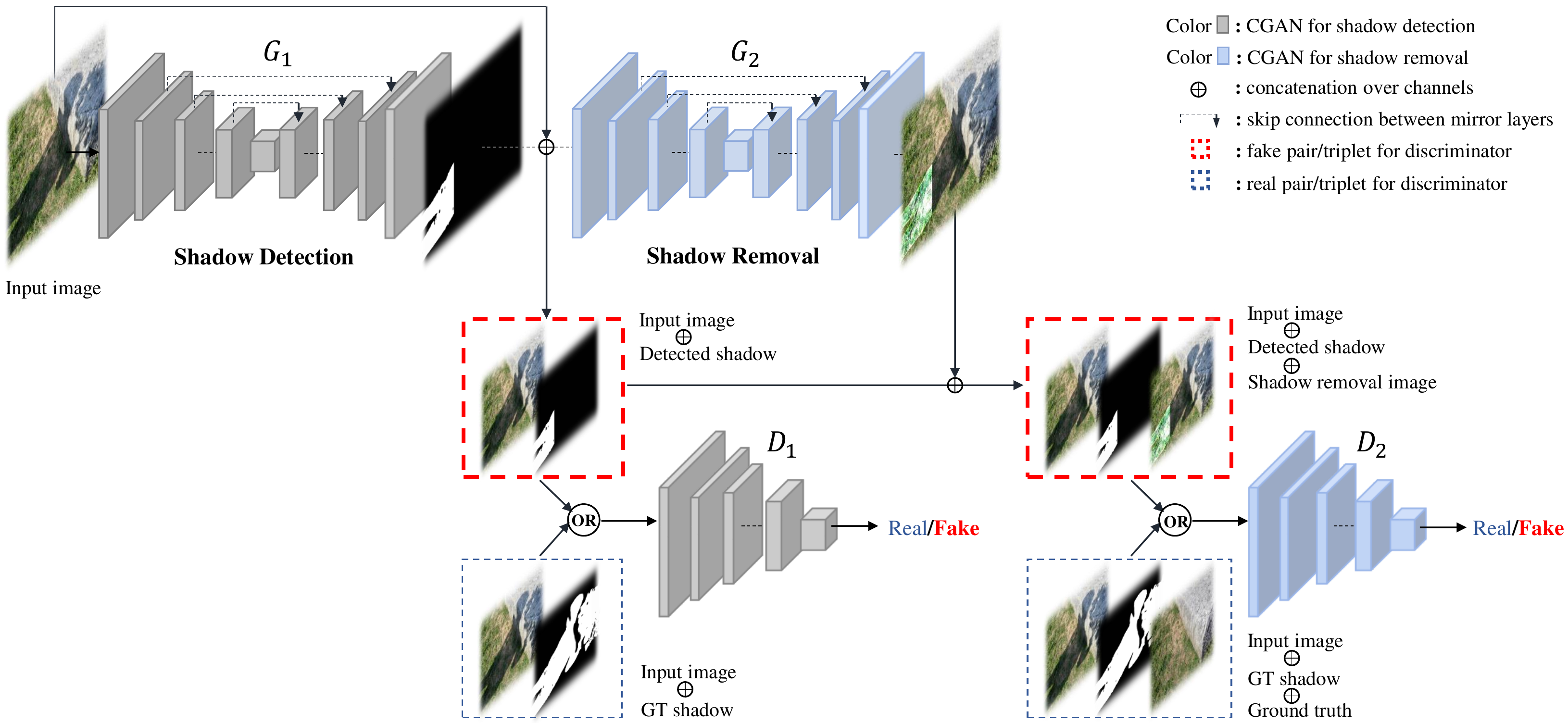}}
	\end{center}
	\vspace{-8pt}
	\caption{The architecture of the proposed ST-CGAN. It consists of two stacked CGANs: one for shadow detection and another for shadow removal, which are marked in different colors. The intermediate outputs are concatenated together as the subsequent components' input. }	
	\label{fig_pipeline_cropped}
	\vspace{-8pt}
\end{figure*}

\textbf{Detection only.} In the history of shadow detection, a series of data-driven statistical learning approaches \cite{huang2011characterizes,lalonde2010detecting,vicente2016noisy,zhu2010learning,khan2014automatic,vicente2015leave} have been proposed. Their main objective is to find the shadow regions, in a form of an image mask that separates shadow and non-shadow areas.

\textbf{Removal only.} A list of approaches \cite{finlayson2006removal,finlayson2009entropy,zhang2015shadow,gryka2015learning,tappen2003recovering,arbel2011shadow,wu2007natural,liu2008texture,qudeshadownet} simply skips the potential information gained from the discovery of shadow regions and directly produces the illumination attenuation effects on the whole image, which is also denoted as a shadow matte \cite{qudeshadownet}, to recover the image with shadows removed naturally.

\textbf{Two stages for removal. } Many of the shadow removal methods \cite{guo2011single,guo2013paired,khan2016automatic,gong2014interactive,vicente2017leave} generally include two \emph{seperated} steps: shadow localization and shadow-free reconstruction by exploiting the intermediate results in the awareness of shadow regions.

It is worth noting that the two targets: shadow mask in detection and shadow-free image in shadow removal, share a fundamental characteristic essentially. As shown in Figure \ref{fig_tasks_cropped}, the shadow mask is posed as a two-binary map that segments the original image into two types of region whereas the shadow removal mainly focuses on one type of that and needs to discover the semantic relationship between the two areas, which indicates the strong correlations and possible mutual benefits between these two tasks.

Besides, most of the previous methods, including shadow detection \cite{huang2011characterizes,lalonde2010detecting,vicente2016noisy,zhu2010learning,khan2014automatic,vicente2015leave} and removal \cite{gong2014interactive,wu2007natural,arbel2011shadow} are heavily based on local region classifications or low-level feature representations, failing to reason about the global scene semantic structure and illumination conditions. Consequently, a most recent study \cite{nguyen2017shadow} in shadow detection introduced a Conditional Generative Adversarial Network (CGAN) \cite{mirza2014conditional} which is proved to be effective for the global consistency. For shadow removal, Qu et al. \cite{qudeshadownet} also proposed a multi-context architecture with an end-to-end manner, which maintained a global view of feature extraction.

Since no existing approaches have explored the joint learning aspect of these two tasks, in this work, we propose a STacked Conditional Generative Adversarial Network (ST-CGAN) framework and aim to tackle shadow detection and shadow removal problems simultaneously in an end-to-end fashion. Besides making full use of the potential mutual promotions between the two tasks, the global perceptions are well preserved through the stacked adversarial components. Further, our design of stacked modules is not only to achieve a multi-task purpose, but also inspired from the connectivity pattern of DenseNet \cite{huang2016densely}, where outputs of all preceding tasks are used as inputs for all subsequent tasks. Specifically, we construct ST-CGAN by stacking two generators along with two discriminators. In Figure \ref{fig_pipeline_cropped}, each generator takes every prior target of tasks (including the input) and stacks them as its input. Similarly, the discriminator attempts to distinguish the concatenation of all the previous tasks' targets from the real corresponding ground-truth pairs or triplets.

Importantly, the design of the proposed stacked components offers a novel perspective for multi-task learning in the literature. Different from the commonly used multi-branch paradigm (e.g., Mask R-CNN \cite{he2017mask}, in which each individual task is assigned with a branch), we stack all the tasks that can not only focus on one task once a time in different stages, but also share mutual improvements through forward/backward information flows. Instead, the multi-branch version aims to learn a shared embedding across tasks by simply aggregating the supervisions from each individual task.

To validate the effectiveness of the proposed framework, we further construct a new large-scale Dataset with Image Shadow Triplets (ISTD) consisting of shadow, shadow mask and shadow-free image to match the demand of multi-task learning. It contains 1870 image triplets under 135 distinct scenarios, in which 1330 is assigned for training whilst 540 is for testing.

Extensive experiments on two large-scale publicly available benchmarks and our newly released dataset show that ST-CGAN performs favorably on both detection and removal aspects, comparing to several state-of-the-art methods. Further, we empirically demonstrate the advantages of our stacked joint formula over the widely used multi-branch version for shadow understanding. To conclude, the main contributions of this work are listed as follows:
\begin{itemize}
	\item It is the first end-to-end framework which jointly learns shadow detection and shadow removal with superior performances on various datasets and on both the two tasks.
	\item A novel STacked Conditional Generative Adversarial Network (ST-CGAN) with a unique stacked joint learning paradigm is proposed to exploit the advantages of multi-task training for shadow understanding.
	\item The first large-scale shadow dataset which contains \emph{image triplets} of shadow, shadow mask and shadow-free image is publicly released.
\end{itemize}

\begin{figure*}[t]
	\setlength{\abovecaptionskip}{1.cm}
	\setlength{\belowcaptionskip}{-1.cm}
	
	\begin{center}
		\setlength{\fboxrule}{0pt}
		\fbox{\includegraphics[width=\textwidth]{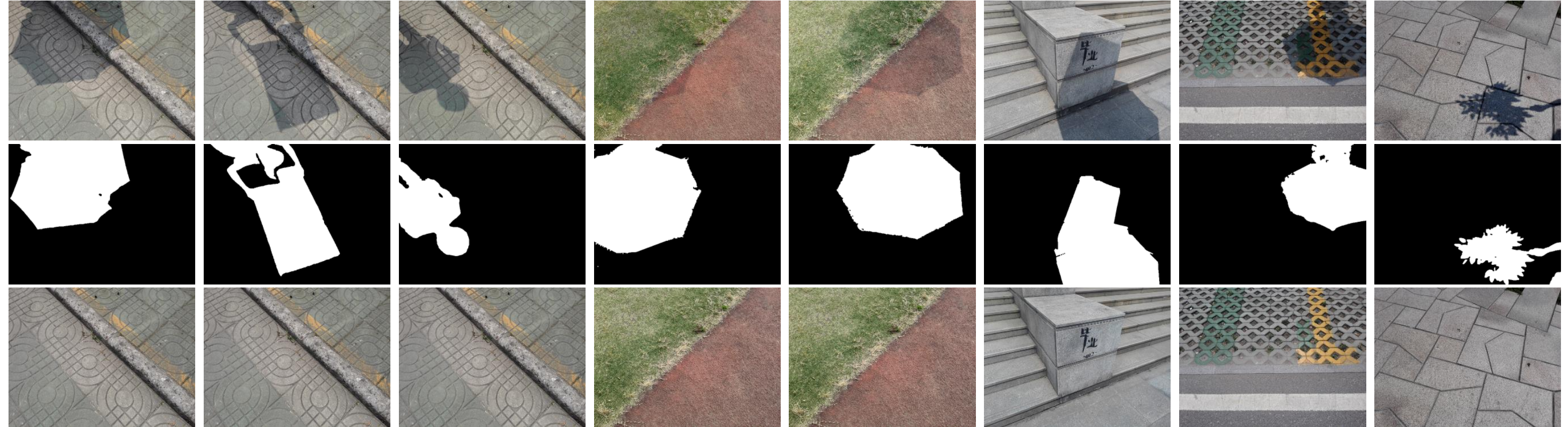}}
	\end{center}
	\vspace{-10pt}
	\caption{An illustration of several shadow, shadow mask and shadow-free image triplets in ISTD. }
	\label{fig_dataset_cropped}
	\vspace{-12pt}
\end{figure*}


\section{Related Work}
\noindent \textbf{Shadow Detection.} To improve the robustness of shadow detection on consumer photographs and web quality images, a series of data-driven approaches \cite{huang2011characterizes,lalonde2010detecting,zhu2010learning} have been taken and been proved to be effective. Recently, Khan et al. \cite{khan2014automatic} first introduced deep Convolutional Neural Networks (CNNs) \cite{simonyan2014very} to automatically learn features for shadow regions/boundaries that significantly outperforms the previous state-of-the-art. A multikernel model for shadow region classification was proposed by Vicente et al. \cite{vicente2015leave} and it is efficiently optimized based on least-squares SVM leave-one-out estimates. More recent work of Vicente et al. \cite{vicente2016noisy} used a stacked CNN with separated steps, including first generating the image level shadow-prior and training a patch-based CNN which produces shadow masks for local patches. Nguyen et al. \cite{nguyen2017shadow} presented the first application of adversarial training for shadow detection and developed a novel conditional GAN architecture with a tunable sensitivity parameter.

\noindent \textbf{Shadow Removal.} Early works are motivated by physical models of illumination and color. For instance, Finlayson et al. \cite{finlayson2009entropy,finlayson2006removal} provide the illumination invariant solutions that work well only on high quality images. Many existing approaches for shadow removal include two steps in general. For the removal part of these two-stage solutions, the shadow is erased either in the gradient domain \cite{finlayson2002removing,mohan2007editing,barron2015shape} or the image intensity domain \cite{arbel2011shadow,guo2011single,guo2013paired,gong2014interactive,khan2016automatic}. On the contrary, a few works \cite{tappen2003recovering,yang2012shadow,qu2015pixel} recover the shadow-free image by intrinsic image decomposition and preclude the need of shadow prediction in an end-to-end manner. However, these methods suffer from altering the colors of the non-shadow regions.  Qu et al. \cite{qudeshadownet} further propose a multi-context architecture which consists of three levels (global localization, appearance modeling and semantic modeling) of embedding networks, to explore shadow removal in an end-to-end and fully automatic framework. 

\noindent \textbf{CGAN and Stacked GAN.} CGANs have achieved impressive results in various image-to-image translation problems, such as image superresolution \cite{ledig2016photo}, image inpainting \cite{pathak2016context}, style transfer \cite{li2016precomputed} and domain adaptation/transfer \cite{isola2016image,zhu2017unpaired,liu2017unsupervised}. The key of CGANs is the introduction of the \emph{adversarial loss} with an informative conditioning variable, that forces the generated images to be with high quality and indistinguishable from real images. Besides, recent researches have proposed some variants of GAN, which mainly explores the stacked scheme of its usage. Zhang et al. \cite{zhang2016stackgan} first put forward the StackGAN to progressively produce photo-realistic image synthesis with considerably high resolution. Huang et al. \cite{huang2016stacked} design a top-down stack of GANs, each learned to generate lower-level representations conditioned on higher-level representations for the purpose of generating more qualified images. Therefore, our proposed stacked form is distinct from all the above relevant versions in essence.

\noindent \textbf{Multi-task Learning.} The learning hypothesis is biased to prefer a shared embedding learnt across multiple tasks. The widely adopted architecture of multi-task formulation is a shared component with multi-branch outputs, each for an individual task. For example, in Mask R-CNN \cite{he2017mask} and MultiNet \cite{teichmann2016multinet}, 3 parallel branches for object classification, bounding-box regression and semantic segmentation respectively are utilized. Misra et al. \cite{Misra_2016_CVPR} propose ``cross-stitch'' unit to learn shared representations from multiple supervisory tasks. In Multi-task Network Cascades\cite{Dai_2016_CVPR}, all tasks share convolutional features, whereas later task also depends the output of a preceding one. 



\section{A new Dataset with Image Shadow Triplets -- ISTD}
Existing publicly available datasets are all limited in the view of multi-task settings. Among them, SBU \cite{vicente2016large} and UCF \cite{zhu2010learning} are prepared for shadow detection only, whilst SRD \cite{qudeshadownet}, UIUC \cite{guo2013paired} and LRSS \cite{gryka2015learning} are constructed for the purpose of shadow removal accordingly.
\begin{table}[h]
	\begin{center}
		
		\vspace{-6pt}
		\resizebox{0.48\textwidth}{!}{ %
			\begin{tabular}{|l|c|l|l|}
				\hline
				Dataset & Amount & Content of Images & Type \\
				\hline \hline
				SRD \cite{qudeshadownet} & 3088 & shadow/shadow-free & pair\\
				\hline
				UIUC \cite{guo2013paired}& 76   & shadow/shadow-free & pair\\
				\hline
				LRSS \cite{gryka2015learning}& 37   & shadow/shadow-free & pair\\
				\hline
				SBU \cite{vicente2016large}& 4727 & shadow/shadow mask & pair\\
				\hline
				UCF \cite{zhu2010learning}& 245  & shadow/shadow mask & pair\\
				\hline\hline
				ISTD (ours) &  1870 & shadow/shadow mask/shadow-free & triplet\\
				\hline
			\end{tabular}
		}
	\end{center}
	\vspace{-5pt}
	\caption{Comparisons with other popular shadow related datasets. Ours is unique in the content and type, whilst being in the same order of magnitude to the most large-scale datasets in amount.}
	\label{tab_datasets}
	\vspace{-5pt}
\end{table}

To facilitate the evaluation of shadow understanding methods, we have constructed a large-scale Dataset with Image Shadow Triplets called \emph{ISTD}\footnote{ISTD dataset is available in https://drive.google.com/file/d/1I0qw-65KBA6np8vIZzO6oeiOvcDBttAY/view?usp=sharing}. It contains 1870 triplets of shadow, shadow mask and shadow-free image under 135 different scenarios. To the best of our knowledge, ISTD is the first large-scale benchmark for simultaneous evaluations of shadow detection and shadow removal. Detailed comparisons with previous popular datasets are listed in Table \ref{tab_datasets}.

In addition, our proposed dataset also contains a variety of properties in the following aspects:
\begin{itemize}
	
	\item \textbf{Illumination:} {Minimized illumination difference} between a shadow image and the shadow-free one is obtained. When constructing the dataset, we pose a camera with a fixed exposure parameter to capture the shadow image, where the shadow is cast by an object. Then the occluder is removed in order to get the corresponding shadow-free image. More evidences are given in the 1st and 3rd row of Figure \ref{fig_dataset_cropped}. 
	
	\item \textbf{Shapes:}  {Various shapes} of shadows are built by different objects, such as umbrellas, boards, persons, twigs and so on. See the 2nd row of Figure \ref{fig_dataset_cropped}.
	
	
	\item \textbf{Scenes:} 135 different types of ground materials, e.g., 6th-8th column in Figure \ref{fig_dataset_cropped}, are utilized to cover as many complex backgrounds and different reflectances as possible. 
\end{itemize}

\begin{table*}[t]
	\begin{center}
		\resizebox{0.98\textwidth}{!}{ %
			\begin{tabular}{|c||c|c|c|c|c|c|c|c|c|c|c|c|c|}
				\hline
				Network & Layer & $\mathrm{Cv_0}$ & $\mathrm{Cv_1}$ & $\mathrm{Cv_2}$ &$\mathrm{Cv_3}$& $\mathrm{Cv_4}$ ($\times 3$) & $\mathrm{Cv_5}$ & $\mathrm{CvT_6}$ & $\mathrm{CvT_7}$ ($\times 3$) & $\mathrm{CvT_8}$ & $\mathrm{CvT_9}$ & $\mathrm{CvT_{10}}$ & $\mathrm{CvT_{11}}$\\
				\hline \hline
				\multirow{5}*{$G_1/G_2$} & \#C\_in&$3/4$&$64$&$128$&$256$&$512$&$512$&$512$&$1024$&$1024$&$512$&$256$&$128$ \\
				& \#C\_out &$64$&$128$&$256$&$512$&$512$&$512$&$512$&$512$&$256$&$128$&$64$&$1/3$ \\
				& before&--&LReLU&LReLU&LReLU&LReLU&LReLU&ReLU&ReLU&ReLU&ReLU&ReLU&ReLU \\
				& after&--&BN&BN&BN&BN&--&BN&BN&BN&BN&BN&Tanh \\
				& link&$\to\mathrm{CvT_{11}}$&$\to\mathrm{CvT_{10}}$&$\to\mathrm{CvT_{9}}$&$\to\mathrm{CvT_{8}}$&$\to\mathrm{CvT_{7}}$&--&--&$\mathrm{Cv_{4}}\to$&$\mathrm{Cv_{3}}\to$&$\mathrm{Cv_{2}}\to$&$\mathrm{Cv_{1}}\to$&$\mathrm{Cv_{0}}\to$ \\
				\hline
			\end{tabular}
		}
	\end{center}	
	\vspace{-5pt}
	\caption{The architecture for generator $G_1/G_2$ of ST-CGAN. $\mathrm{Cv_i}$ means a classic convolutional layer whilst $\mathrm{CvT_i}$ stands for a transposed convolutional layer that upsamples a feature map. $\mathrm{Cv_4}$ ($\times 3$) indicates that the block of $\mathrm{Cv_4}$ is replicated for additional two times, three in total. ``\#C\_in'' and ``\#C\_out'' denote for the amount of input channels and output channels respectively. ``before'' shows the immediate layer before a block and ``after'' gives the subsequent one directly. ``link'' explains the specific connections that lie in U-Net architectures \cite{ronneberger2015u} in which $\to$ decides the direction of connectivity, i.e., $\mathrm{Cv_0}\to\mathrm{CvT_{11}}$ bridges the output of $\mathrm{Cv_0}$ concatenated to the input of $\mathrm{CvT_{11}}$. LReLU is short for Leaky ReLU activation \cite{maas2013rectifier} and BN is a abbreviation of Batch Normalization \cite{ioffe2015batch}.  }
	\label{tab_network_G}
	\vspace{-8pt}
\end{table*}

\section{Proposed Method}
We propose STacked Conditional Generative Adversarial Networks (ST-CGANs), a novel stacked architecture that enables the joint learning for shadow detection and shadow removal, as shown in Figure \ref{fig_pipeline_cropped}. In this section, we first describe the formulations with loss functions, training procedure, and then present the network details of ST-CGAN, followed by a subsequent discussion. 
\subsection{STacked Conditional Generative Adversarial Networks}
Generative Adversarial Networks (GANs) \cite{goodfellow2014generative} consists of two players: a generator $G$ and a discriminator $D$. These two players are competing in a zero-sum game, in which the generator G aims to produce a realistic image given an input $\mathbf{z}$, that is sampled from a certain noise distribution. The discriminator D is forced to classify if a given image is generated by $G$ or it is indeed a real one from the dataset. Hence, the adversarial competition progressively facilitates each other, whilst making $G$'s generation hard for $D$ to differentiate from the real data. Conditional Generative Adversarial Networks (CGANs) \cite{mirza2014conditional} extends GANs by introducing an additional observed information, named conditioning variable, to both the generator $G$ and discriminator $D$. 

Our ST-CGAN consists of two Conditional GANs in which the second one is stacked upon the first.
For the first CGAN of ST-CGAN in Figure \ref{fig_pipeline_cropped}, both the generator $G_1$ and discriminator $D_1$ are conditioned on the input RGB shadow image $\mathbf{x}$. $G_1$ is trained to output the corresponding shadow mask $G_1(\mathbf{z}, \mathbf{x})$, where $\mathbf{z}$ is the random sampled noise vector. We denote the ground truth of shadow mask for $\mathbf{x}$ as $\mathbf{y}$, to which $G_1(\mathbf{z}, \mathbf{x})$ is supposed to be close. As a result, $G_1$ needs to model the distribution $p_{data}(\mathbf{x}, \mathbf{y})$ of the dataset. The objective function for the first CGAN is:
\begin{eqnarray}
\small{ \mathcal{L}_{CGAN_1}(G_1,  D_1)= \mathbf{E}_{\mathbf{x}, \mathbf{y} \sim p_{data}(\mathbf{x}, \mathbf{y})}[\log {D_1(\mathbf{x}, \mathbf{y})}] + }\nonumber\\
\small{
	\mathbf{E}_{\mathbf{x} \sim p_{data}(\mathbf{x}), \mathbf{z} \sim p_{\mathbf{z}}(\mathbf{z})}[\log ({1 - D_1(\mathbf{x}, G_1(\mathbf{z}, \mathbf{x}))})].}
\label{eqa_CGAN1}
\end{eqnarray}


We further eliminate the random variable $\mathbf{z}$ to have a deterministic generator $G_1$ and thus the Equation (\ref{eqa_CGAN1}) is simplified to:
\begin{eqnarray}
\small{ \mathcal{L}_{CGAN_1}(G_1,  D_1)= \mathbf{E}_{\mathbf{x}, \mathbf{y} \sim p_{data}(\mathbf{x}, \mathbf{y})}[\log {D_1(\mathbf{x}, \mathbf{y})}] + }\nonumber\\
\mathbf{E}_{\mathbf{x} \sim p_{data}(\mathbf{x})}[\log ({1 - D_1(\mathbf{x}, G_1(\mathbf{x}))})].
\label{eqa_CGAN1_simple}
\end{eqnarray}

Besides the adversarial loss, the classical data loss is adopted that encourages a straight and accurate regression of the target:
\begin{equation}
\mathcal{L}_{data_1}(G_1)= \mathbf{E}_{\mathbf{x}, \mathbf{y} \sim p_{data}(\mathbf{x}, \mathbf{y})} ||\mathbf{y} - G_1(\mathbf{x})||.
\label{eqa_data_1}
\end{equation}

Further in the second CGAN of Figure \ref{fig_pipeline_cropped}, by applying the similar formulations above, we have:



\begin{table}[t]
	\begin{center}
		\resizebox{0.48\textwidth}{!}{ %
			\begin{tabular}{|c||c|c|c|c|c|c|}
				\hline
				Network & Layer & $\mathrm{Cv_0}$ & $\mathrm{Cv_1}$ & $\mathrm{Cv_2}$ &$\mathrm{Cv_3}$&$\mathrm{Cv_4}$ \\
				\hline \hline
				\multirow{4}*{$D_1/D_2$} & \#C\_in&$4/7$&$64$&$128$&$256$&$512$\\
				& \#C\_out &$64$&$128$&$256$&$512$&$1$ \\
				& before&--&LReLU&LReLU&LReLU&LReLU \\
				& after&--&BN&BN&BN&Sigmoid \\			
				\hline
			\end{tabular}
		}
	\end{center}
	\vspace{-5pt}
	\caption{The architectures for discriminator $D_1/D_2$ of ST-CGAN. Annotations are kept the same with Table \ref{tab_network_G}.}
	\label{tab_network_D}
	\vspace{-8pt}
\end{table}

\begin{equation}
\vspace{+5pt}
\small{\mathcal{L}_{data_2}(G_2 | G_1)= \mathbf{E}_{\mathbf{x}, \mathbf{r} \sim p_{data}(\mathbf{x}, \mathbf{r})} ||\mathbf{r} - G_2(\mathbf{x}, G_1(\mathbf{x}))||,}
\vspace{+15pt}
\end{equation}
\vspace{-15pt}
\begin{equation}
\small{\mathcal{L}_{CGAN_2}(G_2,  D_2 | G_1)= \mathbf{E}_{\mathbf{x}, \mathbf{y}, \mathbf{r} \sim p_{data}(\mathbf{x}, \mathbf{y}, \mathbf{r})}[\log {D_2(\mathbf{x}, \mathbf{y}, \mathbf{r})}]}\nonumber 
\end{equation}
\begin{equation}
+\mathbf{E}_{\mathbf{x} \sim p_{data}(\mathbf{x})}[\log ({1 - \tiny{D_2(\mathbf{x}}, G_1(\mathbf{x}), G_2(\mathbf{x}, G_1(\mathbf{x})))})],
\label{eqa_CGAN2_simple}
\end{equation}

where $\mathbf{r}$ denotes for $\mathbf{x}$'s corresponding shadow-free image and $G_2$ takes a combination of $\mathbf{x}$ and $G_1(\mathbf{x})$ as inputs whereas $D_2$ differentiates the concatenation of outputs from $G_1$ and $G_2$, conditioned on $\mathbf{x}$, from the real pairs. Till this end, we can finally conclude the entire objective for the joint learning task which results in solving a mini-max problem where the optimization aims to find a saddle point:

\vspace{-10pt}
\begin{eqnarray}
\min\limits_{G_1, G_2} \max\limits_{D_1, D_2} \mathcal{L}_{data_1}(G_1) + \lambda_1 \mathcal{L}_{data_2}(G_2 | G_1) + \nonumber\\ 
\lambda_2 \mathcal{L}_{CGAN_1}(G_1,  D_1) + \lambda_3 \mathcal{L}_{CGAN_2}(G_2,  D_2 | G_1).
\end{eqnarray}


It is regarded as a two-player zero-sum game. The first player is a team consisting of two generators ($G_1,G_2$). The second player is a team containing two discriminators ($D_1,D_2$). In order to defeat the second player, the members of the first team are encouraged to produce outputs that are close to their corresponding ground-truths.

\subsection{Network Architecture and Training Details}
\noindent \textbf{Generator.} The generator is inspired by the U-Net architecture \cite{ronneberger2015u}, which is originally designed for biomedical image segmentation. The architecture consists of a contracting path to capture context and a symmetric expanding path that enables precise localization. The detailed structure of $G_1/G_2$, similar to \cite{isola2016image}, is listed in the Table \ref{tab_network_G}.

\noindent \textbf{Discriminator.} For $D_1$, it receives a pair of images as inputs, composed of an original RGB scene image and a shadow mask image that generates 4-channel feature-maps as inputs. The dimensionality of channels increases to 7 for $D_2$ as it accepts an additional shadow-free image. Table \ref{tab_network_D} gives more details of these two discriminators.


\noindent \textbf{Training/Implementation settings.} Our code is based on pytorch \cite{ketkar2017introduction}. We train ST-CGAN with the Adam solver \cite{kingma2014adam} and an alternating gradient update scheme is applied. Specifically, we first adopt a gradient ascent step to update $D_1, D_2$
with $G_1,G_2$ fixed. We then apply a gradient descent step to update $G_1,G_2$
with $D_1, D_2$ fixed. We initialize all the weights of ST-CGAN by sampling from a zero-mean normal distribution with standard deviation 0.2. During training, augmentations are adopted by cropping (image size 286 $\to$ 256) and flipping (horizontally) operations. A practical setting for $\lambda$, where $\lambda_1 = 5, \lambda_2 = 0.1, \lambda_3 = 0.1$, is used. The Binary Cross Entropy (BCE) loss is assigned for the objective of image mask regression and L1 loss is utilized for the shadow-free image reconstruction respectively.

\begin{figure}[t]
	\begin{center}
		\setlength{\fboxrule}{0pt}
		\fbox{\includegraphics[width=0.48\textwidth]{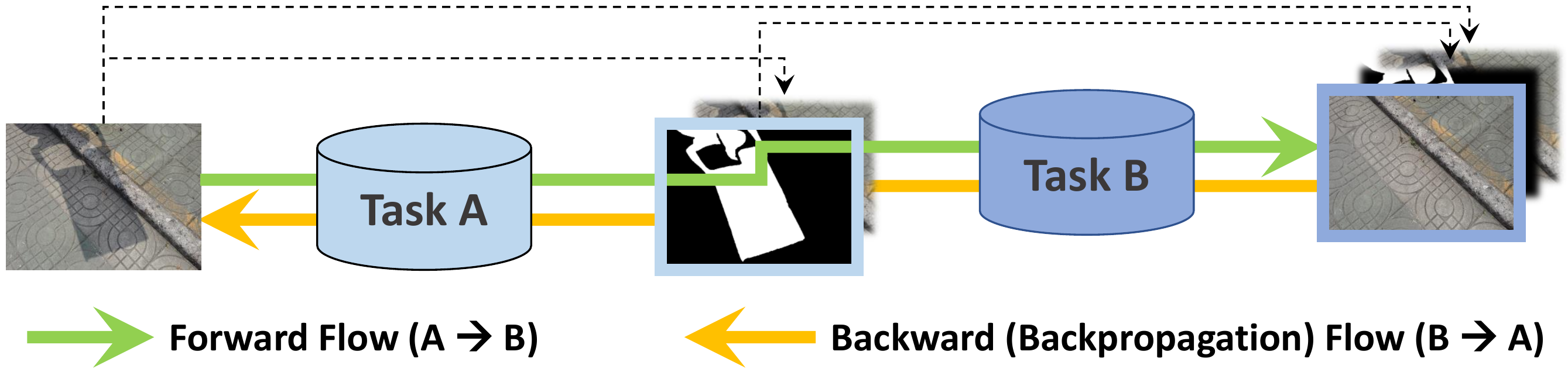}}
	\end{center}
	\vspace{-5pt}
	\caption{An illustration of information flows which indicates the mutual promotions between tasks of the proposed stacked scheme.}
	\label{fig_flow_cropped}
	\vspace{-5pt}
\end{figure}

\begin{table*}[t]
	\begin{center}
		\resizebox{0.82\textwidth}{!}{ %
			\begin{tabular}{|c||c|c|c|c|c|}
				\hline
				Using ISTD Train & Detection Aspects & StackedCNN \cite{vicente2016large} & cGAN \cite{nguyen2017shadow} & scGAN \cite{nguyen2017shadow}  & ours \\
				\hline \hline
				\multirow{3}*{SBU \cite{vicente2016large} (\%)}&Shadow&11.29&24.07&\color{blue}9.1&\color{red}9.02 \\
				& Non-shadow&20.49&\color{red}13.13&17.41&\color{blue}13.66 \\
				& BER&15.94&18.6&\color{blue}13.26&\color{red}{11.34} \\
				\hline \hline
				\multirow{3}*{UCF \cite{zhu2010learning} (\%)}&Shadow &10.56&23.23&\color{blue}9.09&\color{red}8.77 \\
				& Non-shadow&27.58&\color{red}15.61&23.74&\color{blue}23.59 \\
				& BER&18.67&19.42&\color{blue}16.41&\color{red}{16.18} \\
				\hline \hline
				\multirow{3}*{ISTD (\%)}&Shadow &7.96&10.81&\color{blue}3.22&\color{red}2.14 \\
				& Non-shadow&9.23&8.48&\color{blue}6.18&\color{red}5.55 \\
				& BER&8.6&9.64&\color{blue}4.7&\color{red}{3.85} \\
				\hline
			\end{tabular}
		}
	\end{center}
	\vspace{-5pt}
	\caption{Detection with quantitative results using BER, smaller is better. For our proposed architecture, we use image triplets of ISTD training set. These models are tested on three datasets. The best and second best results are marked in {\color{red}{red}} and {\color{blue}{blue}} colors, respectively. }  
	\label{tab_detection_1}
	\vspace{-5pt}
\end{table*}
\begin{table*}[t]
	\begin{center}
		\resizebox{0.82\textwidth}{!}{ %
			\begin{tabular}{|c||c|c|c|c|c|}
				\hline
				Using SBU Train & Detection Aspects & StackedCNN \cite{vicente2016large} & cGAN \cite{nguyen2017shadow} & scGAN \cite{nguyen2017shadow}  & ours \\
				\hline \hline
				\multirow{3}*{SBU \cite{vicente2016large} (\%)}&Shadow&9.6&20.5&\color{blue}7.8&\color{red}3.75 \\
				& Non-shadow&12.5&\color{red}6.9&\color{blue}10.4&12.53 \\
				& BER&11.0&13.6&\color{blue}9.1&\color{red}{8.14} \\
				\hline \hline
				\multirow{3}*{UCF \cite{zhu2010learning} (\%)}&Shadow &9.0&27.06&\color{blue}7.7&\color{red}4.94 \\
				& Non-shadow&17.1&\color{red}10.93&\color{blue}15.3&17.52 \\
				&BER&13.0&18.99&\color{blue}11.5&\color{red}{11.23} \\
				\hline \hline
				\multirow{3}*{ISTD (\%)}&Shadow &11.33&19.93&\color{blue}9.5&\color{red}4.8 \\
				& Non-shadow&9.57&\color{red}4.92&\color{blue}8.46&9.9 \\
				& BER&10.45&12.42&\color{blue}8.98&\color{red}{7.35} \\
				\hline
			\end{tabular}
		}
	\end{center}
	\vspace{-5pt}
	\caption{Detection with quantitative results using BER, smaller is better. For our proposed architecture, we use image pairs of SBU training set together with their roughly generated shadow-free images by Guo et al. \cite{guo2013paired} to form image triplets for training. The best and second best results are marked in {\color{red}{red}} and {\color{blue}{blue}} colors, respectively. }
	\label{tab_detection_2}
	\vspace{-10pt}
\end{table*}

\subsection{Discussion}
\noindent \textbf{The stacked term.} The commonly used form of multi-task learning is the multi-branch version. It aims to learn a shared representation, which is further utilized for each task in parallel. Figure \ref{fig_flow_cropped} implies that our stacked design differs quite a lot from it. We conduct the multi-task learning in such a way that each task can focus on its individual feature embeddings, instead of a shared embedding across tasks, whilst they still enhance each other through the stacked connections, in a form of a forward/backward information flow. The following experiments also confirm the effectiveness of our architecture on the two tasks, compared with the multi-branch one, which can be found in Table \ref{tab_stack_vs_parallel}. 

\noindent \textbf{The {adversarial} term.} Moreover, Conditional GANs (CGANs) are able to effectively enforce higher order consistencies, to learn a joint distribution of image pairs or triplets. This confers an additional advantage to our method, as we implement our basic component to be CGAN and perform a stacked input into the adversarial networks, when compared with nearly most of previous approaches. 

\begin{table*}[t]
	\begin{center}
		\resizebox{0.9\textwidth}{!}{ %
			\begin{tabular}{|c||c|c|c|c|c|c|}
				\hline
				Dataset & Removal aspects & Original & Guo et al. \cite{guo2013paired} & Yang et al. \cite{yang2012shadow} & Gong et al. \cite{gong2014interactive} & ours \\
				\hline \hline
				\multirow{3}*{ISTD }&Shadow & 32.67&18.95&19.82&\color{blue}14.98&\color{red}10.33 \\
				& Non-shadow&\color{red} 6.83&7.46&14.83&7.29&\color{blue}6.93 \\
				& All& 10.97&9.3&15.63&\color{blue}8.53&\color{red}7.47 \\
				\hline
			\end{tabular}
		}
	\end{center}
	\vspace{-5pt}	
	\caption{Removal with quantitative results using RMSE, smaller is better. The original difference between the shadow and shadow-free images is reported in the third column. We perform multi-task training on ISTD and compare it with three state-of-the-art methods. The best and second best results are marked in {\color{red}{red}} and {\color{blue}{blue}} colors, respectively.}
	\label{tab_removal}
	\vspace{-5pt}	
\end{table*}
\begin{table*}[t]
	\begin{center}
		\resizebox{0.90\textwidth}{!}{ %
			\begin{tabular}{|c||c|c|c|c|c|c|c|}
				\hline
				Task Type & Aspects & Ours & Ours (-$D_1$) & Ours (-$D_2$) & Ours (-$G_1$ -$D_1$) & Ours (-$G_2$ -$D_2$)\\
				\hline \hline
				\multirow{3}{*}{Removal} &Shadow & \color{red}10.33 &\color{blue}10.36& 10.38 &  12.12 & -- \\
				&Non-shadow&\color{red}6.93&\color{blue}6.96&7.03&7.45&--\\
				&All &\color{red}7.47&\color{blue}7.51&7.56&8.19&-- \\
				\hline \hline
				\multirow{3}*{Detection (\%)}&Shadow &\color{red} 2.14&2.62&\color{blue}2.49& -- & 3.4\\
				& Non-shadow& \color{blue}5.55&6.18&6.03&-- &\color{red}5.1 \\
				& BER&\color{red}3.85 &4.4&4.26&-- &\color{blue}4.25\\
				\hline
			\end{tabular}
		}
	\end{center}
	\vspace{-5pt}		
	\caption{Component analysis of ST-CGAN on ISTD by using RMSE for removal and BER for detection, smaller is better. The metrics related to shadow and non-shadow part are also provided. The best and second best results are marked in {\color{red}{red}} and {\color{blue}{blue}} colors, respectively.}
	\label{tab_component}
	\vspace{-10pt}	
\end{table*}

\section{Experiments}
To comprehensively evaluate the performance of our proposed method, we perform extensive experiments on a variety of datasets and evaluate ST-CGAN in both detection and removal measures, respectively.
\subsection{Datasets}
We mainly utilize two large-scale publicly available datasets\footnote{Note that we do not include the large-scale \textbf{SRD} dataset in this work because it is currently unavailable for the authors' \cite{qudeshadownet} personal reasons.} including SBU \cite{vicente2016large} and UCF \cite{zhu2010learning}, along with our newly collected dataset ISTD.



\noindent \textbf{SBU \cite{vicente2016large}} has 4727 pairs of shadow and shadow mask image. Among them, 4089 pairs are for training and the rest is for testing.

\noindent \textbf{UCF \cite{zhu2010learning}} has 245 shadow and shadow mask pairs in total, which are all used for testing in the following experiments.

\noindent \textbf{ISTD} is our new released dataset consisting of 1870 triplets, which is suitable for multi-task training. It is randomly divided into 1330 for training and 540 for testing.

\vspace{-5pt}
\subsection{Compared Methods and Metrics}
\noindent \textbf{For detection part,} we compare ST-CGAN with the state-of-the-art StackedCNN \cite{vicente2016large}, cGAN \cite{nguyen2017shadow} and scGAN \cite{nguyen2017shadow}. To evaluate the shadow detection performance quantitatively, we follow the commonly used terms \cite{nguyen2017shadow} to compare the provided ground-truth masks and the predicted ones with the main evaluation metric, which is called Balance Error Rate (BER):
\begin{equation}
\mathrm{BER} = 1 - \frac{1}{2}(\frac{TP}{TP + FN} + \frac{TN}{TN + FP}),
\end{equation}
along with separated per pixel error rates per class (shadow and non-shadow).

\noindent \textbf{For removal part,} we use the publicly available source codes \cite{guo2013paired,yang2012shadow,gong2014interactive} as our baselines. In order to perform a quantitative comparison, we follow \cite{guo2013paired,qudeshadownet} and use the root mean square error (RMSE) in LAB color space between the ground truth shadow-free image and the recovered image as measurement, and then evaluate the results on the whole image as well as shadow and non-shadow regions separately.

\subsection{Detection Evaluation}
For detection, we utilize the cross-dataset shadow detection schedule, similar in \cite{nguyen2017shadow}, to evaluate our method. We first train our proposed ST-CGAN on the ISTD training set. The evaluations are thus conducted on three datasets with three state-of-the-art approaches in Table \ref{tab_detection_1}. As can be seen, ST-CGAN outperforms StackedCNN and cGAN by a large margin. In terms of BER, we obtain a significant 14.4\% error reduction on SBU and 18.1\% on ISTD respectively, compared to scGAN. 

Next, we switch the training set to SBU's training data. Considering our framework requires image triplets that SBU cannot offer, we make an additional pre-processing step. In order to get the corresponding shadow-free image, we use the shadow removal code \cite{guo2013paired} to generate them as coarse labels. We also test these trained models on the three datasets. Despite the inaccurate shadow-free ground-truths, our proposed framework still significantly improves the overall performances. Specifically, on the SBU test set, ST-CGAN achieves an obvious improvement with 10.5\% error reduction of BER over the previous best record from scGAN.

In Figure \ref{fig_select_cropped}, we demonstrate the comparisons of the detection results qualitatively. As shown in Figure \ref{fig_select_cropped} (a) and \ref{fig_select_cropped} (b), ST-CGAN is not easily fooled by the lower brightness area of the scene, comparing to cGAN and scGAN. Our method is also precise in detecting shadows cast on bright areas such as the line mark in Figure \ref{fig_select_cropped} (c) and \ref{fig_select_cropped} (d). The proposed ST-CGAN is able to detect more fine-grained shadow details (e.g., shadow of leaves) than other methods, as shown in Figure \ref{fig_select_cropped} (e) and \ref{fig_select_cropped} (f).

\subsection{Removal Evaluation}
For removal, we compare our proposed ST-CGAN with the three state-of-the-art methods on ISTD dataset, as shown in Table \ref{tab_removal}. The RMSE values are reported. We evaluate the performance of different methods on the shadow regions, non-shadow regions, and the whole image. The proposed ST-CGAN achieves the best performance among all the compared methods by a large margin. Notably, the error of non-shadow region is very close to the original one, which indicates its strong ability to distinguish the non-shadow part of an image. The advantage of removal also partially comes from the joint learning scheme, where the well-trained detection block provides more clear clues of shadow and shadow-free areas. 

\begin{figure*}[t]
	\vspace{-15pt}
	\begin{center}
		\setlength{\fboxrule}{0pt}
		\fbox{\includegraphics[width=\textwidth]{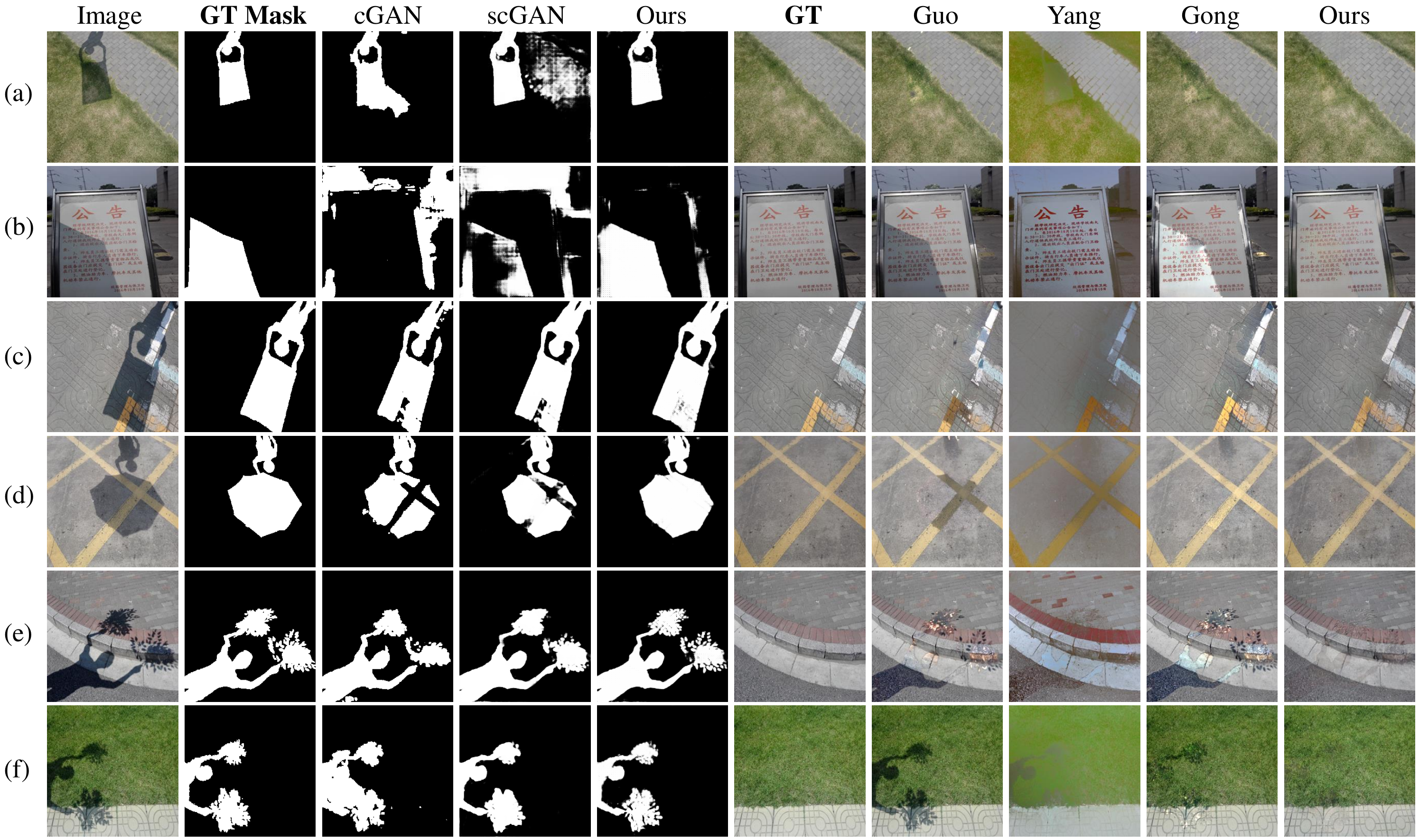}}
	\end{center}	
	\vspace{-10pt}
	\caption{Comparison of shadow detection and removal results of different methods on ISTD dataset. Note that our proposed ST-CGAN simultaneously produces the detection and removal results, whilst others are either for shadow detection or for shadow removal.}
	\label{fig_select_cropped}
	\vspace{-10pt}
\end{figure*}

We also demonstrate the comparisons of the removal results. As shown in Figure \ref{fig_select_cropped}, although Yang \cite{yang2012shadow} can recover shadow-free image, it alters the colors of both shadow and nonshadow regions. Guo \cite{guo2011single} and Gong \cite{gong2014interactive} fail to detect shadow accurately, thus both of their predictions are incomplete especially in shadow regions. Moreover, due to the difficulty of determining the environmental illuminations and global consistency, all the compared baseline models produce unsatisfactory results on the semantic regions.



\vspace{-1pt}
\subsection{Component Analysis of ST-CGAN}
To illustrate the effects of different components of ST-CGAN, we make a series of ablation experiments by progressively removing different parts of it. According to both the removal and the detection performances in Table \ref{tab_component}, we find that each individual component is necessary and indispensable for the final excellent predictions. Moreover, the last two columns of Table \ref{tab_component} also demonstrate that without the stacked joint learning, a single module consisting of one generator and one discriminator performs worse consistently. It further implies the effectiveness of our multi-task architecture on both shadow detection and shadow removal.

\begin{table}[t]
	\vspace{-1pt}
	\begin{center}
		\resizebox{0.48\textwidth}{!}{ %
			\begin{tabular}{|c||c|c|c|}
				\hline
				Task Type & Aspects & Multi-branch & Ours \\
				\hline \hline
				\multirow{3}{*}{Removal} &Shadow & 11.54 & \textbf{10.33}\\
				&Non-shadow& 7.13 & \textbf{6.93}\\
				&All & 7.84 & \textbf{7.47} \\
				\hline \hline
				\multirow{3}{*}{Detection (\%)} &Shadow & 2.34 & \textbf{2.14}\\
				&Non-shadow& 7.2 & \textbf{5.55}\\
				&BER&4.77&  \textbf{3.85}\\
				\hline
			\end{tabular}
		}
	\end{center}
	\vspace{-5pt}		
	\caption{Comparisons between stacked learning (ours) and multi-branch learning with removal and detection results on ISTD dataset. }
	\label{tab_stack_vs_parallel}
	\vspace{-12pt}	
\end{table}
\subsection{Stacked Joint \emph{vs.} Multi-branch Learning}
\begin{figure}[t]
	\vspace{-2pt}
	\begin{center}
		\setlength{\fboxrule}{0pt}
		\fbox{\includegraphics[width=0.46\textwidth]{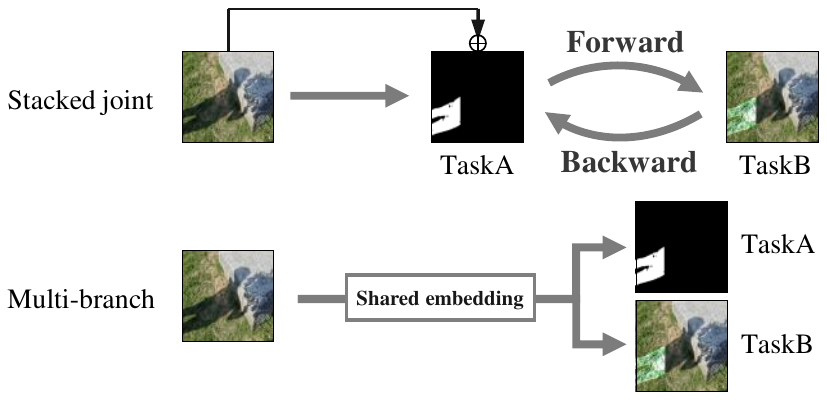}}
	\end{center}
	\vspace{-10pt}	
	\caption{Illustrations of our stacked joint learning (top) and common multi-branch learning (bottom).}
	\label{fig_parallel_cropped}
	\vspace{-10pt}	
\end{figure}

We further modify our body architecture into a multi-branch version, where each branch is designed for one task respectively. Therefore, the framework aims to learn a shared embedding which is supervised by two tasks, as shown in the bottom of Figure \ref{fig_parallel_cropped}. For a clear explanation, the illustration of comparisons between ours and the multi-branch one is also given.  With all other training settings fixed, we fairly compare our proposed ST-CGAN with the multi-branch version quantitatively on the measurements of both detection and removal on ISTD dataset. Table \ref{tab_stack_vs_parallel} reports that our stacked joint learning paradigm consistently outperforms the multi-branch version in every single aspect of the metrics.

\section{Conclusion}
In this paper, we have proposed STacked Conditional Generative Adversarial Network (ST-CGAN) to jointly learn shadow detection and shadow removal. Our framework has at least four unique advantages as follows: 1) it is the first end-to-end approach that tackles shadow detection and shadow removal simultaneously; 2) we design a novel stacked mode, which densely connects all the tasks in the purpose of multi-task learning, that proves its effectiveness and suggests the future extension on other types of multiple tasks; 3) the stacked adversarial components are able to preserve the global scene characteristics hierarchically, thus it leads to a fine-grained and natural recovery of shadow-free images; 4) ST-CGAN consistently improves the overall performances on both the detection and removal of shadows. Moreover, as an additional contribution, we publicly release the first large-scale dataset which contains shadow, shadow mask and shadow-free image triplets.



{\small
\bibliographystyle{ieee}
\bibliography{egbib}
}

\end{document}